\documentclass[conference]{IEEEtran}
\usepackage{times}
\usepackage{eso-pic,url,xcolor}

\usepackage[numbers]{natbib}
\usepackage{multicol}
\usepackage[bookmarks=true]{hyperref}
\usepackage{amsmath,amsthm,amssymb,mathtools,bbm}
\usepackage{subcaption} 
\usepackage{graphicx}
\usepackage{algorithm, algorithmic}
\usepackage{xcolor}
\usepackage{cancel}

\theoremstyle{plain}

\theoremstyle{definition}

\usepackage{mathtools}
\usepackage{url}

\newcommand\scalemath[2]{\scalebox{#1}{\mbox{\ensuremath{\displaystyle #2}}}}

\begin{document}
\newcommand{\confyear}{2026}

\AddToShipoutPicture*{\AtPageUpperLeft{%
\begin{minipage}{\paperwidth}
\vspace{2.8cm}
\centering
\ttfamily\small
\color[rgb]{.5,.5,.5}%
Accepted for presentation at \emph{Robotics: Science and Systems (RSS)}, \confyear.\\
This is the authors' preprint version.
\end{minipage}}}

\title{ELVIS: Ensemble-Calibrated Latent Imagination for Long-Horizon Visual MPC}

\IEEEoverridecommandlockouts
\author{
Yurui Du$^{1,3}$, Pinhao Song$^{2,3}$, Yutong Hu$^{2,3}$, and Renaud Detry$^{1,2,3}$%
\thanks{$^{1}$Yurui Du and Renaud Detry are with KU Leuven, Dept. Electrical Engineering, Research unit Processing Speech and Images, B-3000 Leuven, Belgium.
E-mail: \texttt{\{yurui.du, renaud.detry\}@kuleuven.be}.}%
\thanks{$^{2}$Pinhao Song, Yutong Hu, and Renaud Detry are with KU Leuven, Dept. Mechanical Engineering, Research unit Robotics, Automation and Mechatronics, B-3000 Leuven, Belgium.
E-mail: \texttt{\{pinhao.song, yutong.hu\}@kuleuven.be}.}%
\thanks{$^{3}$Yurui Du, Pinhao Song, Yutong Hu, and Renaud Detry are with Flanders Make at KU Leuven, B-3000 Leuven, Belgium.}%
}

\maketitle

\begingroup
\renewcommand{\thefootnote}{}
\footnotetext{
\begin{center}
\includegraphics[height=1.2cm]{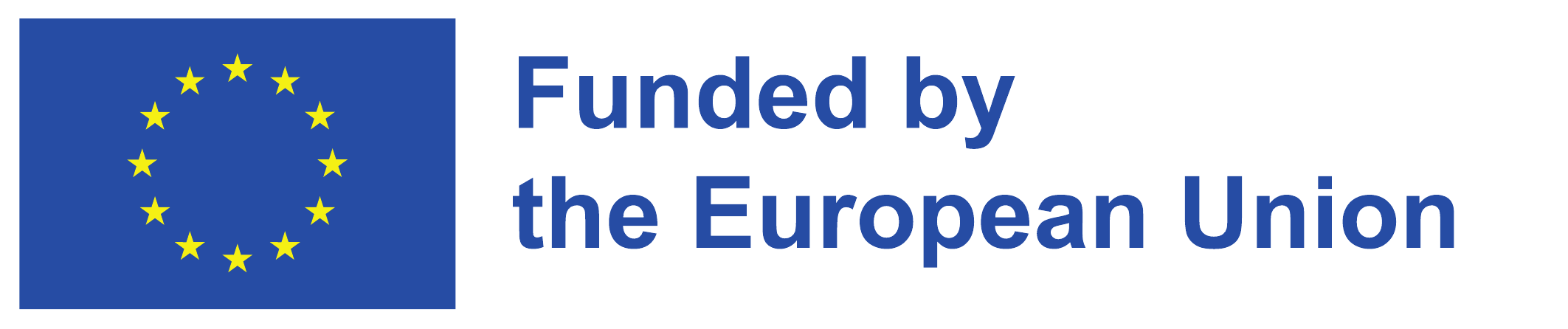}
\end{center}

This work was supported by Interne Fondsen KU Leuven/Internal Funds KU Leuven and the European Union (\url{robetarme-project.eu}). Views and opinions expressed are however those of the authors only and do not necessarily reflect those of the European Union or HADEA. Neither the European Union nor the granting authority can be held responsible for them.

}
\endgroup

\begin{abstract}
A central challenge of visual control with model-based reinforcement learning (RL) is \emph{reliable long-horizon planning}: Long rollouts with learned latent dynamics exhibit branching futures and multi-modal action-value distributions. In addition, compounding model error amplified by visual occlusions make deep imagination brittle. We present \textbf{ELVIS}, a latent model predictive controller (MPC) designed to make long-horizon planning practical. ELVIS plans in a Dreamer-style recurrent state space model (RSSM) and replaces standard unimodal model predictive path integral (MPPI) with a \emph{Gaussian-mixture MPPI} that maintains multiple coherent hypotheses over long horizons, avoiding mode averaging under branching rollouts. In parallel, ELVIS stabilizes deep imagination with a \emph{shared uncertainty-aware $\lambda_t$-return}: an ensemble of latent critics defines an upper-confidence-bound (UCB) score that gates a time-varying $\lambda_t$, adaptively trading off bootstrapping versus look-ahead to limit compounding error during planning. The same return is used both to train an actor-critic prior from imagined rollouts and to score candidate trajectories inside GMM-MPPI, aligning RL objectives with the planner’s long-horizon optimization. On fourteen DeepMind Control Suite visual tasks, ELVIS establishes state-of-the-art performance compared with TD-MPC2 \cite{hansen2024tdmpc2} and DreamerV3 \cite{Hafner2025DreamerV3Nature}. Finally, ELVIS transfers zero-shot to a real-world sand spraying task with severe occlusions, improving surface-quality metrics and demonstrating robustness beyond simulation. Our code is publicly available at \url{https://elvis-mpc.github.io}.
\end{abstract}

\IEEEpeerreviewmaketitle

\section{Introduction}

Model-based RL has emerged as a promising alternative to model-free methods, achieving state-of-the-art sample efficiency by learning a world model and reducing reliance on exhaustive trial-and-error. Typically, one first learns a predictive dynamics model via supervised training, then leverages it to either (i) generate imaginary rollouts that augment policy/value learning \citep{hafner2019learning,hafner2019dream,Hafner2025DreamerV3Nature}, or (ii) support an MPC that optimizes predicted trajectories online to provide world-informed control decisions \cite{hansen2024tdmpc2,hansen2022temporal,nagabandi2018neural}. Short-horizon planning/imagined rollouts provide effective “look-ahead” that reduces wasteful real-world exploration and improves data efficiency. Finally, when MPC is warm-started or guided by a learned policy prior, the planner can reliably choose actions that outperform the raw policy, improving sampling efficiency and stability \citep{wang2025bootstrapped}.

Despite these advances, MPC in visual control is often limited to short receding horizons (e.g., 3--5 steps), largely because longer-horizon latent rollouts are both computationally expensive and increasingly error-prone. As the rollout horizon grows, model errors compound and imagined returns become less reliable; this problem becomes especially acute under partial observability, where occlusions reduce visual evidence and force the planner to rely more heavily on latent imagination~\cite{kayalibay2023filteraware}. Nonetheless, longer horizons should, in principle, substantially improve sample efficiency, provided planning can \emph{reason about and react to uncertainty} so that unreliable imagined rollouts do not dominate decisions.
A second limitation is the Markov assumption underlying many MPC instantiations. RL-based methods often heuristically \emph{stack a short history of frames} as a Markov proxy—e.g., DQN stacks 4 Atari frames—then learn on this proxy state \citep{Mnih2015}. While this can mitigate partial observability in benign settings, it quickly breaks under occlusions, viewpoint shifts, or distractors \citep{Stone2021DCS}, leading to state aliasing and drift over longer rollouts. In such regimes, agents need \emph{memory/belief state} rather than short stacks \citep{Igl2018}.

To address these limitations, we propose ELVIS, an uncertainty-aware, memory-augmented MPC framework for visual model-based RL under partial observability. Two existing model-based RL lines are especially relevant to this setting. Dreamer-style agents~\citep{Hafner2025DreamerV3Nature} use recurrent state-space models (RSSMs) to maintain latent belief under partial observability, making them a natural foundation for visual control with occlusion. In parallel, TD-MPC~\citep{hansen2022temporal} and TD-MPC2~\citep{hansen2024tdmpc2} show that online trajectory optimization in the latent space of a learned world model can yield strong control performance. ELVIS combines these strengths: it uses a Dreamer-style recurrent latent belief for memory under occlusion, while extending TD-MPC2-style latent planning to longer horizons through multimodal proposals and uncertainty-aware return modulation.

Concretely, we employ an RSSM~\citep{hafner2019learning} to capture aleatoric uncertainty in high-dimensional observations through its stochastic latent state, while modeling epistemic uncertainty via an ensemble of value networks~\citep{sunrise21g}. We instantiate informative data collection while guarding against brittle long-horizon predictions using uncertainty-based soft truncation that continuously trades off short-horizon bootstrapping against long-horizon rollout returns. This yields a unified interface between recurrent memory, long-horizon planning, and uncertainty-aware control.

Our paper makes the following contributions:

1) Adaptive-horizon, memory-augmented MPC for visual RL. A framework that adapts the effective return horizon online under controlled model error, using real-time confidence thresholds and compute–performance trade-offs, built on a recurrent latent world model for partial observability.



2) Uncertainty-regulated exploration and exploitation. A unified exploration-exploitation scheme bounded by soft truncation, yielding uncertainty-aware actions and more reliable long-horizon plans.

3) We evaluate ELVIS on both DeepMind visual control benchmarks and challenging real-world tasks featuring extreme partial occlusions. 

Collectively, these elements produce a practical recipe for long-horizon, uncertainty-aware planning in visual model-based RL. Across various settings, our method consistently improves data efficiency, robustness under partial observability, and zero-shot transfer performance compared to state-of-the-art model-based RL baselines \cite{Hafner2025DreamerV3Nature, hansen2024tdmpc2}.

\section{Related Work}

\subsection{Sampling-Based MPC and Multimodal Latent Planning}
Model Predictive Path Integral (MPPI) control provides a sampling-based framework for online trajectory optimization in nonlinear systems~\cite{williams16mppi}, and TD-MPC brings this idea into learned latent spaces for continuous control~\cite{hansen2022temporal,hansen2024tdmpc2}. ELVIS follows this latent-MPC lineage but targets a long-horizon regime where unimodal TD-MPC2-style proposals can become brittle: branching imagined futures may cause a single Gaussian to average over incompatible action-sequence hypotheses. We address this with multimodal MPPI in latent imagination. Existing multimodal MPPI methods in reactive task-and-motion planning and collision avoidance handle modes induced by high-level plan alternatives or maneuver choices~\cite{zhang24mmppi,bertipaglia25mmppi}, while Bayesian MBRL methods such as PaETS also argue for maintaining multimodal trajectory hypotheses~\cite{okada20vimpc}. In contrast, ELVIS studies multimodal proposals inside a recurrent latent world model and couples them to a shared uncertainty-aware return used by both imagined learning and online planning.


\subsection{Uncertainty Quantification in Model-Based RL}
Uncertainty-aware model-based RL has been studied from several angles: PETS propagates epistemic uncertainty in learned dynamics~\cite{chua18pets}, Plan2Explore uses world-model disagreement as an intrinsic reward~\cite{sekar2020plan2explore}, and recent latent-space methods use epistemic uncertainty for out-of-distribution detection and safety filtering~\cite{seo25unisafe}. ELVIS differs in the role assigned to uncertainty. Rather than using it primarily for dynamics propagation, exploration, or safety triggering, we use critic uncertainty to gate a time-varying $\lambda_t$, softly modulating imagined returns. This aligns the uncertainty signal with the return propagation, while avoiding the cost of maintaining uncertainty estimates inside the RSSM dynamics.

\subsection{Optimism, Value-Guided Planning, and Trust-Aware Rollouts}
ELVIS is also related to ensemble-based optimism and trust-aware rollout control. SUNRISE uses ensemble-UCB style optimism for exploration~\cite{sunrise21g}, and POLO combines online trajectory optimization with learned value functions and uncertainty-aware exploration~\cite{lowrey18polo}. However, neither SUNRISE nor POLO is designed to address when long-horizon imagined rollouts should be trusted during learning. MBPO and MACURA instead regulate model usage by scheduling, shortening, or adapting rollout lengths based on model uncertainty~\cite{janner2019mbpo,frauenknecht2024macura}. ELVIS uses critic uncertainty for a different purpose: it modulates how strongly distant imagined returns propagate inside a fixed horizon. The resulting time-varying $\lambda_t$ reduces the influence of uncertain distant returns during both imagined actor-critic learning and MPPI scoring. This distinction is computationally relevant in our online setting, where variable rollout lengths can introduce system overhead such as more frequent recompilation, whereas a fixed-horizon planner with soft return modulation preserves a stable execution path.

\section{Preliminaries}
This section summarizes the recurrent state-space model (RSSM) used throughout the paper and clarifies how it supports control under partial observability. We first describe RSSM as a latent-variable sequence model trained by variational inference, then explain how the learned latent dynamics support downstream return optimization in imagination.

\begin{figure}[t]
  \centering
  \includegraphics[width=\columnwidth]{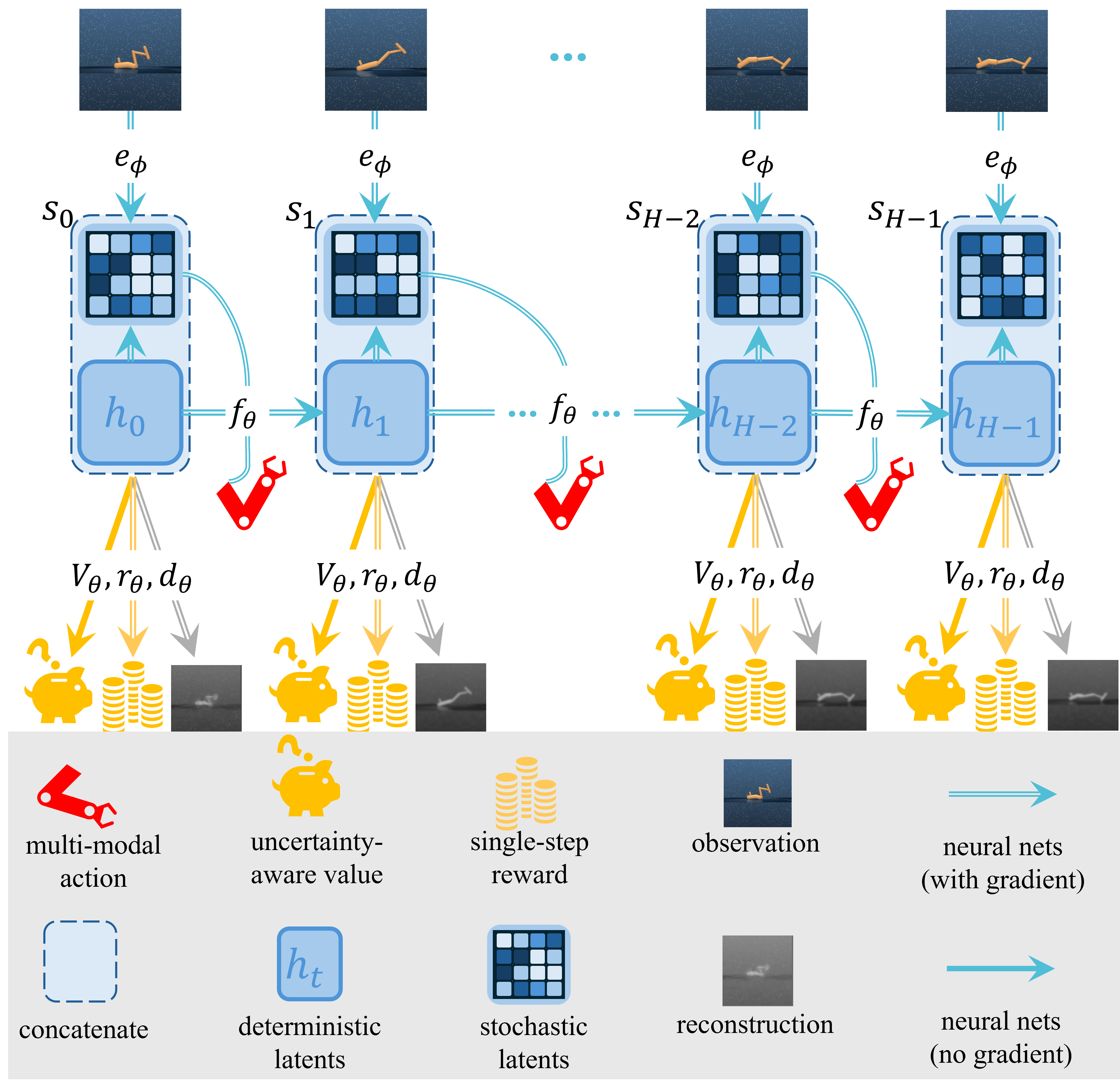}
  \caption{\textbf{RSSM world model learning under partial observability.}
An encoder infers stochastic latents $z_t$ from observations conditioned on a recurrent memory state $h_t$, which is updated by a deterministic transition given actions. The learned prior predicts future latents, while decoders reconstruct observations and rewards, yielding a compact belief state $\hat{s}_t=(h_t,z_t)$ for latent imagination and downstream planning.}
  \label{fig:dream-world}
\end{figure}

We consider partially observable control from trajectories
\(
\tau=\{o_t,a_t,r_t\}_{t=0}^{T}
\),
where \(o_t\) is the observation, \(a_t\) the action, and \(r_t\) the reward. The control objective is to maximize the discounted return
\begin{equation}
\label{eq:rl-objective}
J(\pi)
=
\mathbb{E}_{p(\tau)}
\!\left[
\sum_{t=0}^{T}\gamma^t r_t
\right].
\end{equation}
Following Dreamer-style latent imagination~\cite{Hafner2025DreamerV3Nature}, RSSM represents the filtered belief at time \(t\) by
\(
\hat s_t=(h_t,z_t)
\),
where \(h_t\) is a deterministic memory state summarizing observation history and \(z_t\) is a stochastic latent capturing aleatoric uncertainty.

Given actions \(a_{0:T-1}\), RSSM learning seeks to maximize the conditional evidence
\begin{equation}
\label{eq:rssm-evidence}
\max_{\theta,\phi}\;
\log p_\theta(o_{0:T},r_{0:T}\mid a_{0:T-1}),
\end{equation}
where \(\theta\) denotes the generative model parameters and \(\phi\) the encoder parameters. The RSSM consists of a posterior encoder, a deterministic transition, a latent prior, an observation decoder, and a reward model:
\begin{equation}
\label{eq:rssm-components}
\begin{aligned}
z_t &\sim e_\phi(z_t\mid h_t,o_t), \\
h_{t+1} &= f_\theta(h_t,z_t,a_t)=F_\theta(z_{:t},a_{:t})=F_{\theta}^{:t}, \\
z_{t+1} &\sim p_\theta(z_{t+1}\mid h_{t+1}), \\
o_t &\sim d_\theta(o_t\mid h_t,z_t), \\
r_t &\sim r_\theta(r_t\mid h_t,z_t).
\end{aligned}
\end{equation}
To incorporate the deterministic memory update into the probabilistic model, we write it as a Dirac transition. The resulting joint model is
\begin{equation}
\label{eq:rssm-joint}
\begin{aligned}
&p_\theta(o_{0:T},r_{0:T},z_{0:T},h_{0:T}\mid a_{0:T-1}) \\
&=
p(h_0)\,
p_\theta(z_0\mid h_0)\,
d_\theta(o_0\mid h_0,z_0)\,
r_\theta(r_0\mid h_0,z_0) \\
&\times
\prod_{t=0}^{T-1}
\delta\!\big(h_{t+1}-f_\theta(h_t,z_t,a_t)\big)\,
p_\theta(z_{t+1}\mid h_{t+1}) \\
&
\times\,
d_\theta(o_{t+1}\mid h_{t+1},z_{t+1})\,
r_\theta(r_{t+1}\mid h_{t+1},z_{t+1}),
\end{aligned}
\end{equation}
while the filtering posterior is
\begin{equation}
\label{eq:rssm-posterior}
\begin{aligned}
&q_\phi(z_{0:T},h_{0:T}\mid o_{0:T},a_{0:T-1}) =
p(h_0)\,
e_\phi(z_0\mid h_0,o_0) \\
&\times
\prod_{t=0}^{T-1}
\delta\!\big(h_{t+1}-f_\theta(h_t,z_t,a_t)\big)\,
e_\phi(z_{t+1}\mid h_{t+1},o_{t+1}).
\end{aligned}
\end{equation}
Applying the standard variational identity gives
\begin{equation}
\label{eq:rssm-elbo-derivation}
\begin{aligned}
&\log p_\theta(o_{0:T},r_{0:T}\mid a_{0:T-1}) \\
&=
\log \!\iint\!
q_\phi
\frac{
p_\theta(o_{0:T},r_{0:T},z_{0:T},h_{0:T}\mid a_{0:T-1})
}{
q_\phi(z_{0:T},h_{0:T}\mid o_{0:T},a_{0:T-1})
}
\,\text{d}z_{0:H}\,\text{d}h_{0:H} \\
&\ge
\mathbb{E}_{q_\phi}
\!\left[
\log p_\theta(o_{0:T},r_{0:T},z_{0:T},h_{0:T}\mid a_{0:T-1})
-
\log q_\phi
\right] \\
&=: \mathrm{ELBO}_{\mathrm{RSSM}} .
\end{aligned}
\end{equation}
Substituting \eqref{eq:rssm-joint} and \eqref{eq:rssm-posterior}, the Dirac terms cancel, yielding
\begin{equation}
\label{eq:rssm-elbo}
\begin{aligned}
&\mathrm{ELBO}_{\mathrm{RSSM}}
=
\mathbb{E}_{q_\phi}
\Bigg[
\sum_{t=0}^{T}
\log d_\theta(o_t\mid h_t,z_t) \\
&
+\log r_\theta(r_t\mid h_t,z_t) 
+\log p_\theta(z_t\mid h_t)
-\log e_\phi(z_t\mid h_t,o_t)
\Bigg] \\
&=
\sum_{t=0}^{T}
\mathbb{E}_{q_\phi}
\Big[
\log d_\theta(o_t\mid F_{\theta}^{:t-1},z_t)
+
\log r_\theta(r_t\mid F_{\theta}^{:t-1},z_t)
\Big] \\
&-
\sum_{t=0}^{T}
\Big[
\mathrm{KL}\!\left(
e_\phi(\cdot\mid F_{\theta}^{:t-1},o_t)
\,\|\,p_\theta(\cdot\mid F_{\theta}^{:t-1})
\right)
\Big].
\end{aligned}
\end{equation}
Thus, RSSM learning maximizes \(\mathrm{ELBO}_{\mathrm{RSSM}}\), which jointly encourages accurate observation reconstruction, reward prediction, and consistency between the filtered posterior and the latent prior.

After training, the posterior encoder maps the current history to a filtered latent belief \(\hat s_t=(h_t,z_t)\), which serves as the starting state for imagination. Future latent rollouts then replace posterior updates with prior transitions under the learned RSSM, and predicted rewards are accumulated to approximate the return in Eq.~\ref{eq:rl-objective}. Dreamer-style methods use these imagined trajectories for actor-critic updates. In this paper, we focus on further return maximization under latent imagination using long-horizon MPC.

\begin{figure}[t]
  \centering
  \includegraphics[width=\columnwidth]{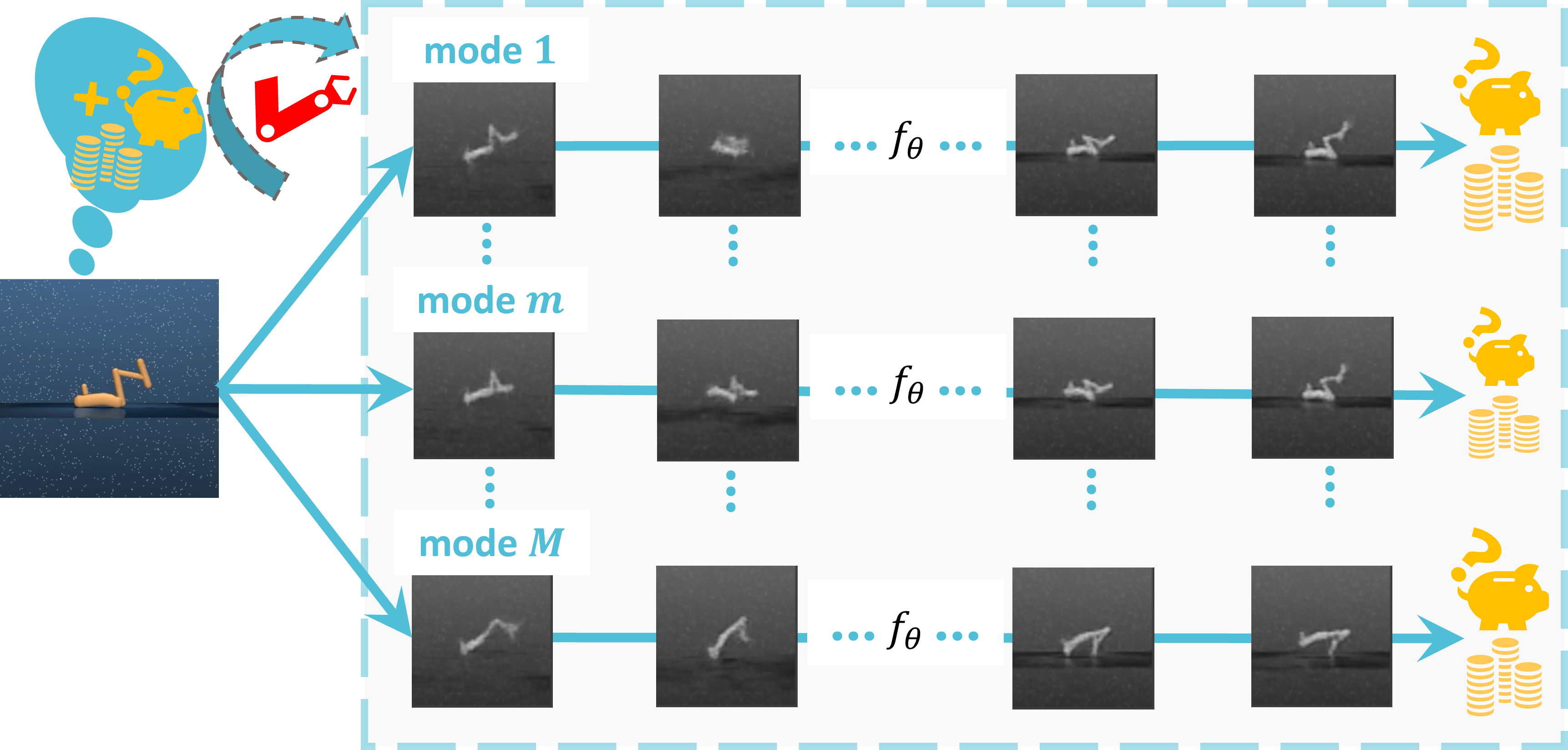}
  \caption{\textbf{GMM-based long-horizon MPPI for multimodal trajectory distributions.}
  Over long horizons, sampled rollouts diverge and form multiple distinct high-reward modes.
  We fit a Gaussian Mixture Model (GMM) to the trajectory (or action-sequence) samples to capture this multimodality,
  then perform MPPI-style weighting and control extraction \emph{per mode} before aggregating into a single action.
  This reduces mode collapse compared to a single-Gaussian update and improves robustness under partial observability.}
  \label{fig:gmm-mppi}
\end{figure}
\section{Methods}
Our goal is long-horizon visual control under partial occlusion. Starting from the RSSM belief state $\hat{s}_t=(h_t,z_t)$, we make long-horizon MPC practical by addressing two coupled failure modes: (i) \emph{trajectory branching} over long imagination horizons, which makes unimodal MPPI updates collapse; and (ii) \emph{compounding epistemic uncertainty} from learned dynamics and missing observations, which makes deep rollouts unreliable unless planning and learning know when to truncate. We therefore introduce (1) \textbf{GMM-MPPI} to represent multi-modal long-horizon trajectory distributions, and (2) a \textbf{shared uncertainty-aware return} (UCB-gated $\lambda_t$) used for both imagined actor--critic learning and MPPI trajectory scoring, enabling an uncertainty-driven trade between short-horizon bootstrapping and long-horizon rollout returns (\emph{soft truncation}) that is particularly effective under occlusions.

\subsection{GMM-MPPI for Multi-Modal Long-Horizon Planning}
\label{sec:gmm-mppi}

At each environment step, we optimize an $H$-length action sequence in latent space using MPPI conditioned on the current belief $\hat{s}_k=(h_k,z_k)$. Standard MPPI maintains a unimodal (typically Gaussian) proposal over action sequences and updates its mean after reweighting samples by their predicted return. Prior stochastic-planning / MPC work has shown that optimal trajectory or action distributions can be multimodal, making unimodal Gaussian proposals limiting in some settings~\cite{okada20vimpc,honda2024svgmppi}. In long-horizon latent planning under partial observability, this issue becomes especially acute: branching imagined futures can cause initially similar action sequences to lead to increasingly separated future states, while several such futures remain competitive in return. As illustrated in Fig.~\ref{fig:gmm-mppi}, a single-Gaussian proposal then becomes a poor approximation of the high-return distribution, averaging across incompatible hypotheses and producing conservative ``mean'' sequences that do not correspond to any coherent future. This motivates richer proposal families for planning.

To preserve multiple plausible long-term hypotheses, we maintain \emph{$M$ parallel Gaussian proposals} (modes) over action sequences:
\begin{equation}
\label{eq:gmm-proposal-rewrite}
q_m(a_{0:H-1}) \;=\; \mathcal N\!\big(a_{0:H-1};\mu_m,\Sigma_m\big),
\qquad m=1,\dots,M,
\end{equation}
where $\mu_m=\{\mu_{m,t}\}_{t=0}^{H-1}$ and $\Sigma_m=\{\Sigma_{m,t}\}_{t=0}^{H-1}$ are time-indexed moments (often diagonal). For each mode $m$, we draw $K$ candidates, $\{a^{(k)}_{0:H-1}\}_{k=1}^{K}\sim q_m(\cdot)$, roll each sequence out under the learned RSSM latent dynamics starting from $\hat{s}_k$, and assign a scalar score $G^{(k)}_m$ using the uncertainty-aware return in Sec.~\ref{sec:uncertainty-return} (Eq.~\ref{eq:lambda-return}). To normalize scores consistently across modes, we compute the best score over \emph{all} sampled rollouts,
\begin{equation}
\label{eq:global-best}
G^\star \;=\; \max_{m\in\{1,\dots,M\},\,k\in\{1,\dots,K\}} G^{(k)}_m.
\end{equation}
We then compute \emph{mode-wise} weights using a relative-to-best softmax (normalized \emph{within} each mode),
\begin{equation}
\label{eq:modewise-softmax-global}
w^{(k)}_{m}
\;=\;
\frac{\exp\!\left(\frac{1}{\tau}\,\frac{G^{(k)}_m}{G^\star+\delta}\right)}
{\sum_{j=1}^{K}\exp\!\left(\frac{1}{\tau}\,\frac{G^{(j)}_m}{G^\star+\delta}\right)},
\qquad k=1,\dots,K,
\end{equation}
where $\tau>0$ is a temperature and $\delta>0$ is a small constant for numerical stability. Each mode is updated by weighted moment matching:
\begin{align}
\label{eq:gmm-mean-rewrite}
\mu_{m,t}
&\leftarrow
\sum_{k=1}^{K} w^{(k)}_{m}\,a^{(k)}_t,
\qquad t=0,\dots,H-1,\\
\label{eq:gmm-cov-rewrite}
\Sigma_{m,t}
&\leftarrow
\sum_{k=1}^{K} w^{(k)}_{m}\,
\big(a^{(k)}_t-\mu_{m,t}\big)\big(a^{(k)}_t-\mu_{m,t}\big)^\top
\;+\;\epsilon I,
\end{align}
with a small regularizer $\epsilon I$. Unlike a unimodal proposal, these parallel updates preserve multiple coherent hypotheses under branching latent trajectories and stabilize long-horizon optimization.

\paragraph{Policy--random initialization and warm-start.}
We warm-start planning across time by shifting each mean sequence $\mu_m$ forward (receding horizon). To diversify hypotheses within a planning call, we initialize each mode $m$ using a distinct policy--random mixing ratio $\alpha_m\in[0,1]$:
\begin{equation}
\label{eq:policy-random-init}
\mu_m \leftarrow \alpha_m a^{\pi}_{0:H-1} + (1-\alpha_m)a^{\mathrm{rand}}_{0:H-1},
\end{equation}
where $a^{\pi}_{0:H-1}\sim \pi_\psi(\cdot\mid \hat{s}_k)$ is a proposal from the learned actor prior and $a^{\mathrm{rand}}_{0:H-1}$ is a random sequence (e.g., Gaussian noise within action bounds). After $L$ MPPI iterations, we select the highest-scoring rollout across modes (Eq.~\ref{eq:global-best}) and execute the first action of its corresponding mean sequence.

\subsection{Uncertainty-aware learning and planning via UCB-gated $\lambda_t$}
\label{sec:uncertainty-return}
Long-horizon imagination is only useful if we can control compounding model error. This is particularly critical for \emph{visual occlusions}: when some observations are missing, the belief must be propagated by the RSSM prior, increasing epistemic uncertainty and making deep rollouts unreliable unless the controller can decide \emph{where the effective horizon should end}. We address this with a shared uncertainty-aware return used consistently for: (i) learning an imagined actor--critic prior and (ii) scoring MPPI rollouts.

\paragraph{Ensemble-UCB uncertainty}
We maintain an ensemble of latent critics $\{V_i(\hat{s})\}_{i=1}^M$ and compute, along imagined rollouts,
\begin{equation}
\label{eq:ens-moments}
\begin{aligned}
\mu_t &:= \frac{1}{M}\sum_{i=1}^M V_i(\hat{s}_t),\\
\sigma_t &:= \sqrt{\frac{1}{M-1}\sum_{i=1}^M \big(V_i(\hat{s}_t)-\mu_t\big)^2 }.
\end{aligned}
\end{equation}
These define an optimism-flavored importance score, where $\beta$ is used to deliver the exploration-exploitation tradeoff.
\begin{equation}
\label{eq:ucb}
\mathrm{UCB}(\hat{s}_t) \;:=\; \mu_t + \beta\,\sigma_t .
\end{equation}
UCB is high when a predicted future is both \emph{valuable} (large mean) and \emph{epistemically informative} (large dispersion), which is precisely the regime we wish to reach under occlusion-driven ambiguity via long-horizon exploration.

\paragraph{Soft truncation with time-varying $\lambda_t$}
Our uncertainty-based soft truncation is built upon standard $\lambda$ return, except that we map UCB to a time-dependent $\lambda_t$ to determine how far return information should propagate through imagination:
\begin{equation}
\label{eq:lambda-ucb}
\begin{aligned}
\lambda_t
&=\lambda_{\max}
-(\lambda_{\max}-\lambda_{\min})\,
\mathrm{norm}\!\big(\mathrm{UCB}(\hat{s}_t)\big),\\
&\text{s.t.}\quad 0 \le \lambda_{\min} \le \lambda_{\max} \le 1,
\end{aligned}
\end{equation}
so that \emph{high UCB} induces \emph{smaller} $\lambda_t$ (more bootstrapping, shorter effective horizon), while \emph{low UCB} induces \emph{larger} $\lambda_t$ (deeper look-ahead). This yields an uncertainty-triggered soft truncation: once a high-UCB future is reached, we reduce reliance on deeper (noisier) imagination; if the rollout remains uninformative, we keep propagating rewards further to search for better futures. Here $\mathrm{norm}(\cdot)\in[0,1]$ denotes a running-statistics normalization (e.g., exponential moving mean/variance with clipping) computed over recent imagined rollouts, following standard stabilization practices used in world-model and latent-MPC agents~\citep{Hafner2025DreamerV3Nature,hansen2024tdmpc2}.

\begin{figure}[t]
  \centering
  \includegraphics[width=\columnwidth]{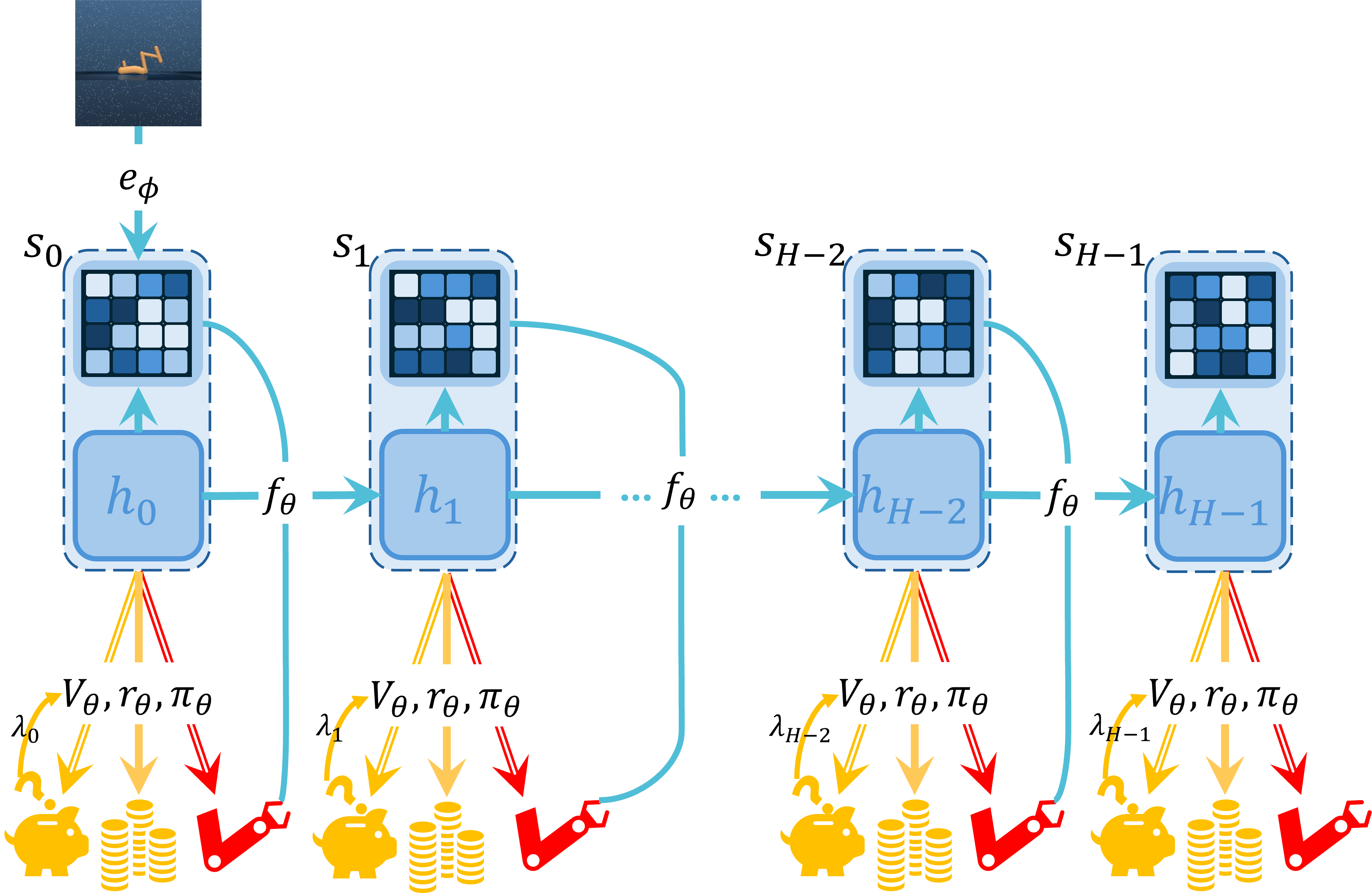}
  \caption{\textbf{Imaginary TD learning with UCB-gated $\lambda$-returns.}
  We train actor--critic priors from RSSM-imagined rollouts using an ensemble-UCB score to set a time-varying $\lambda_t$ in the $\lambda$-return targets. High-UCB states induce smaller $\lambda_t$ (greater bootstrapping), while low-UCB states induce larger $\lambda_t$ (deeper look-ahead), yielding stable yet exploratory value and policy learning for MPPI warm-starting.}
  \label{fig:ucb-lambda}
\end{figure}

\paragraph{Shared return for imagined TD learning and MPPI scoring}
We define the time-varying $\lambda$-return recursively as
\begin{equation}
\label{eq:lambda-return}
\begin{aligned}
G_t \;&=\; \hat r_t + \gamma\Big((1-\lambda_t)\,\mu_{t+1} + \lambda_t\,G_{t+1}\Big),\\
G_{H-1} \;&=\; \mu_{H-1},
\end{aligned}
\end{equation}
where $\hat r_t$ is the RSSM reward prediction. Crucially, we use the \emph{same} UCB-gated return for \emph{both} learning and planning: $G_0$ serves as the TD target for training the critic ensemble and an actor prior $\pi_\theta(a\mid\hat{s})$ as shown in Fig. \ref{fig:ucb-lambda}. The learned actor then warm-starts MPPI by initializing the action-sequence mean (and/or injecting policy-sampled candidates), which improves sample efficiency and robustness under partial occlusions. This design explicitly aligns with the view that policy learning and MPC should collaborate: RL provides a strong prior that guides sampling-based planning, while MPC provides an online improvement operator that continually refines the policy's proposals~\citep{wang2025bootstrapped}.

We summarize our method, ELVIS, in Alg. \ref{alg:elvis_legacy}.

\begin{algorithm}[t]
\caption{ELVIS: Online learning and inference with GMM-MPPI and UCB-gated $\lambda_t$-returns}
\label{alg:elvis_legacy}
\small
\begin{algorithmic}[1]
\STATE Initialize RSSM world model and posterior belief update (Dreamer-style)~\citep{Hafner2025DreamerV3Nature}
\STATE Initialize actor prior $\pi_\theta$, critic ensemble $\{V_i\}_{i=1}^{E}$, replay buffer $\mathcal D$
\STATE Initialize per-mode moments $\{\mu_m,\Sigma_m\}_{m=1}^{M}$
\STATE
\STATE \textbf{Function} \textsc{Plan}$(\hat s_k)$
\STATE \quad Warm-start all modes via receding-horizon shift
\STATE \quad Initialize each mode via Eq.~\ref{eq:policy-random-init}
\STATE \quad \textbf{for} $\ell=1$ \textbf{to} $L$ \textbf{do}
\STATE \qquad \textbf{for} $m=1$ \textbf{to} $M$ \textbf{do}
\STATE \qquad\quad Sample $K$ sequences from Eq.~\ref{eq:gmm-proposal-rewrite}
\STATE \qquad\quad Score rollouts using Eqs.~\ref{eq:ens-moments}--\ref{eq:lambda-return}
\STATE \qquad \textbf{end for}
\STATE \qquad Compute global best $G^\star$ via Eq.~\ref{eq:global-best}
\STATE \qquad \textbf{for} $m=1$ \textbf{to} $M$ \textbf{do}
\STATE \qquad\quad Compute weights via Eq.~\ref{eq:modewise-softmax-global}
\STATE \qquad\quad Update $(\mu_m,\Sigma_m)$ via Eqs.~\ref{eq:gmm-mean-rewrite}--\ref{eq:gmm-cov-rewrite}
\STATE \qquad \textbf{end for}
\STATE \quad \textbf{end for}
\STATE \quad Return first action of the best mode (Eq.~\ref{eq:global-best})
\STATE
\FOR{environment steps $k=1,2,\dots$}
  \STATE Update belief $\hat s_k$ using RSSM posterior~\citep{Hafner2025DreamerV3Nature}
  \STATE $a_k \leftarrow$ \textsc{Plan}$(\hat s_k)$; execute $a_k$; store transition in $\mathcal D$
  \STATE Update RSSM (standard world-model learning)~\citep{Hafner2025DreamerV3Nature}
  \STATE Update critics and actor prior using shared return (Eq.~\ref{eq:lambda-return})
\ENDFOR
\end{algorithmic}
\end{algorithm}

\begin{figure*}[t]
  \centering
  \includegraphics[width=\textwidth]{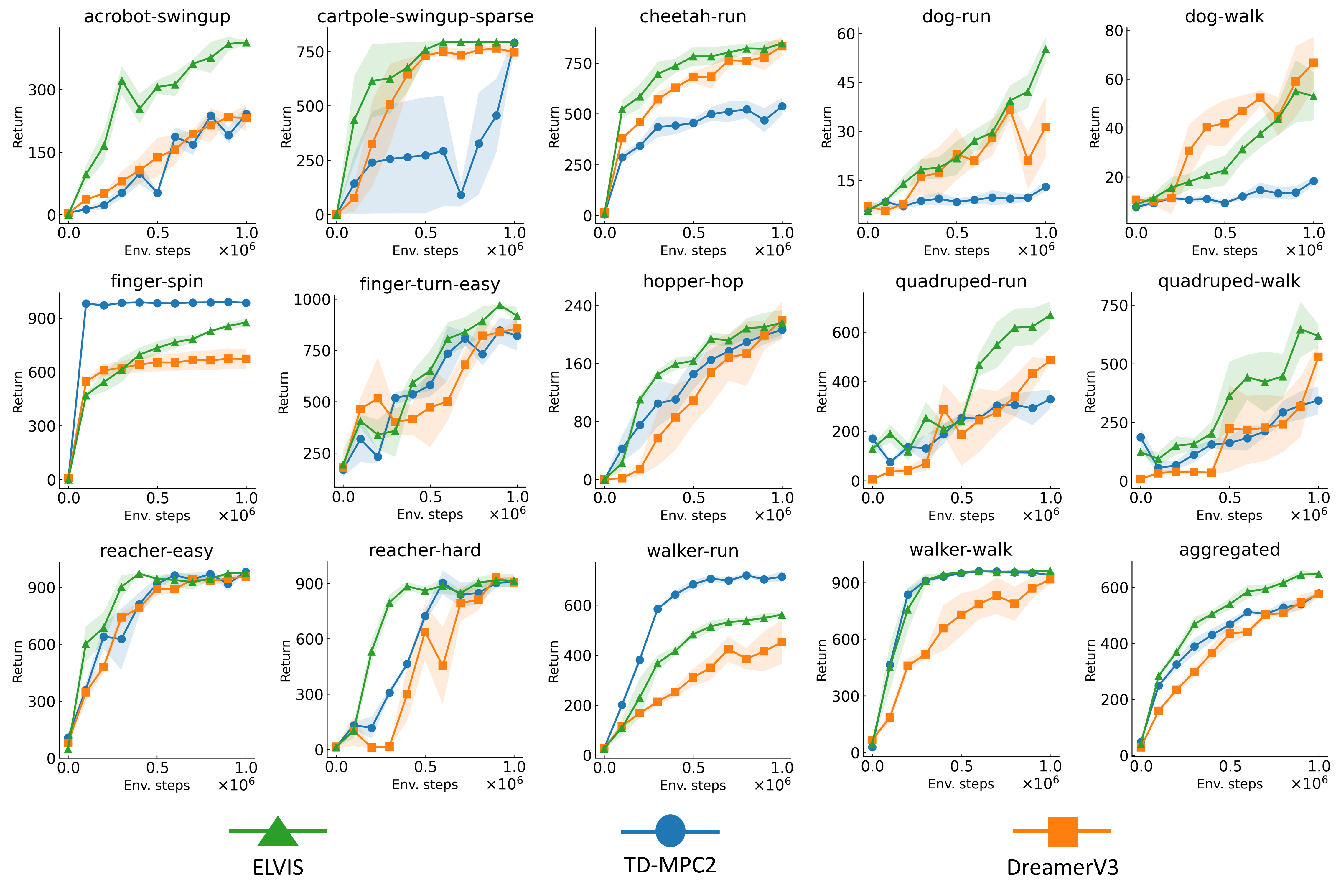}
  \caption{\textbf{DMC visual control learning curves.}
  Per-task learning curves on 14 DeepMind Control (DMC) visual control benchmarks, together with an aggregated score that reports the mean episodic return averaged across all 14 tasks at each environment step. Shaded regions denote 95\% confidence intervals over 5 random seeds. ELVIS achieves the strongest overall performance, ranking first or second on every task.}
  \label{fig:dmc-all}
\end{figure*}

\begin{figure}[t]
  \centering
  \includegraphics[width=\columnwidth]{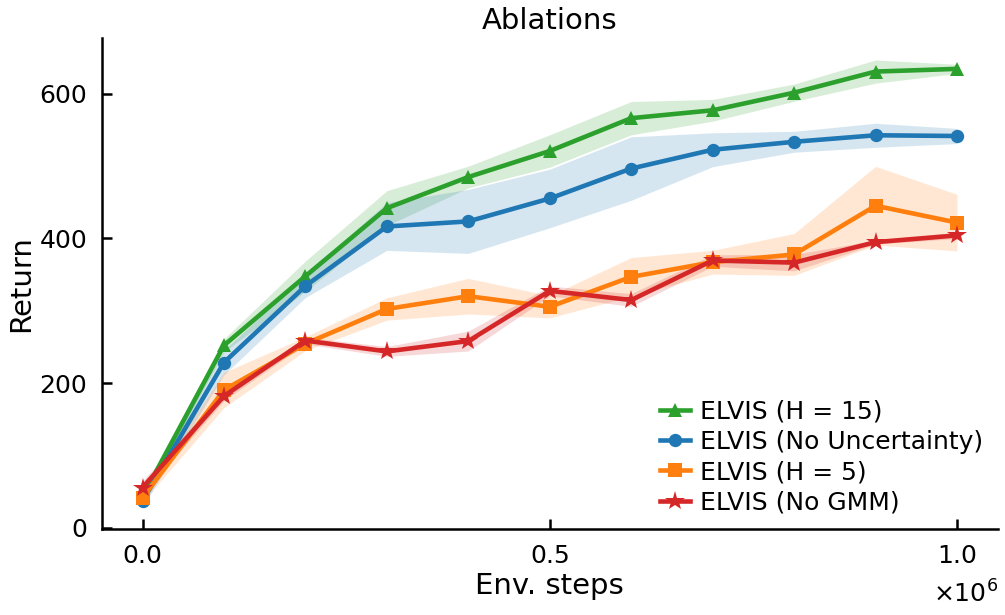}
  \caption{\textbf{Ablations of ELVIS.}
  Aggregated learning curves on the same 14 DMC visual control tasks, where the score is the mean episodic return averaged across tasks at each environment step. Shaded regions denote 95\% confidence intervals over 5 random seeds. Removing either GMM, uncertainty awareness or long-horizon planning degrades sample efficiency and final performance, indicating that all three components contribute and interact to produce ELVIS's gains.}
  \label{fig:dmc-ablation}
\end{figure}
\section{Experiments}

We structure experiments to answer two primary questions. 
\textbf{Q1: Does long-horizon, uncertainty-aware MPC improve performance on visual control?} 
To address this, we evaluate \textbf{TD-MPC2}~\citep{hansen2024tdmpc2}, \textbf{DreamerV3}~\citep{Hafner2025DreamerV3Nature}, and \textbf{ELVIS} on 14 DeepMind Control Suite (DMC) vision tasks. This comparison isolates complementary strengths: TD-MPC2 provides a strong online planner based on frame stacking instead of explicit recurrent memory, whereas DreamerV3 provides recurrent world-model learning without online planning. We follow standard protocols (10 evaluation episodes every 100k environment steps, 5 seeds). 

We answer this in two steps. First, we report results under standard DMC visual control tasks, directly testing whether long-horizon planning and uncertainty awareness are effective for visual control in Sec.~\ref{sec:dmc_visual}. Second, we perform ablations on DMC to isolate the roles of \emph{planning horizon} and \emph{uncertainty-aware truncation} by sweeping horizons and removing uncertainty mechanisms (fixed-$\lambda$/no-UCB variants), quantifying how long-horizon gains depend on uncertainty control in Sec.~\ref{sec:dmc_visual}.

\textbf{Q2: Under extreme partial occlusions in the real world, which components are essential for robust performance and sim-to-real transfer?}
To address this question, we evaluate \textbf{zero-shot sim-to-real transfer} on a real-world sand-spraying task featuring extreme occlusions and sensory noise~\citep{du2025arepo}. This setting emphasizes the necessity of \emph{explicit memory} and tests whether \emph{uncertainty-aware planning} improves robustness beyond simulation. We use the surface-quality metrics and protocol of~\citep{du2025arepo} and compare against the same baselines in Sec.~\ref{sec:shotcrete}).


\subsection{Visual Control on DMC}
\label{sec:dmc_visual}


Across 14 visual control learning tasks, \textbf{ELVIS establishes state-of-the-art performance}, improving both sample efficiency and final return relative to the baselines in the majority of environments, as shown in Fig.~\ref{fig:dmc-all}. Notably, ELVIS is consistently competitive: it ranks \emph{first or second} across tasks and is never the lowest-performing method among the three. This pattern is important because ELVIS was motivated by the need to handle more challenging visual control settings with severe partial observability and occlusions (Section~\ref{sec:shotcrete}); yet the results here show that its design choices also translate to \emph{standard} visual control benchmarks. In other words, the components introduced for robustness under partially observable settings, uncertainty awareness and long-horizon model-based planning, do not trade off performance in easier settings, but instead improve data efficiency and stabilize performance broadly. This suggests a promising direction for improving model-based RL: explicitly accounting for predictive uncertainty while planning over longer horizons can yield benefits even when observations are complete. 

To isolate the key drivers of performance, we conduct controlled ablations on the same DMC suite and report \emph{aggregated} performance (mean return averaged across the 14 tasks at each evaluation step) in Fig.~\ref{fig:dmc-ablation}. First, varying the MPC horizon shows that \textbf{long-horizon planning is crucial}: $H\!=\!15$ \textbf{substantially outperforms} $H\!=\!5$, indicating that deeper foresight is beneficial when paired with a learned world model. Second, replacing our uncertainty-based $\lambda$-return truncation with a non-uncertainty variant (e.g., fixed-$\lambda$ / no-UCB) causes a \textbf{large drop} in aggregated score, showing that \textbf{uncertainty awareness is necessary to control compounding model error}. Third, replacing the proposed GMM action proposal with a \textbf{unimodal Gaussian} (\textbf{No GMM}) also lowers performance, supporting our claim that long-horizon latent planning benefits from \textbf{multimodal action proposals} rather than a single Gaussian.



\begin{figure}[t]
    \centering
    \begin{minipage}[t]{0.48\textwidth} 
        \centering
        \begin{subfigure}[t]{0.23\textwidth}
            \centering
            \includegraphics[width=\textwidth]{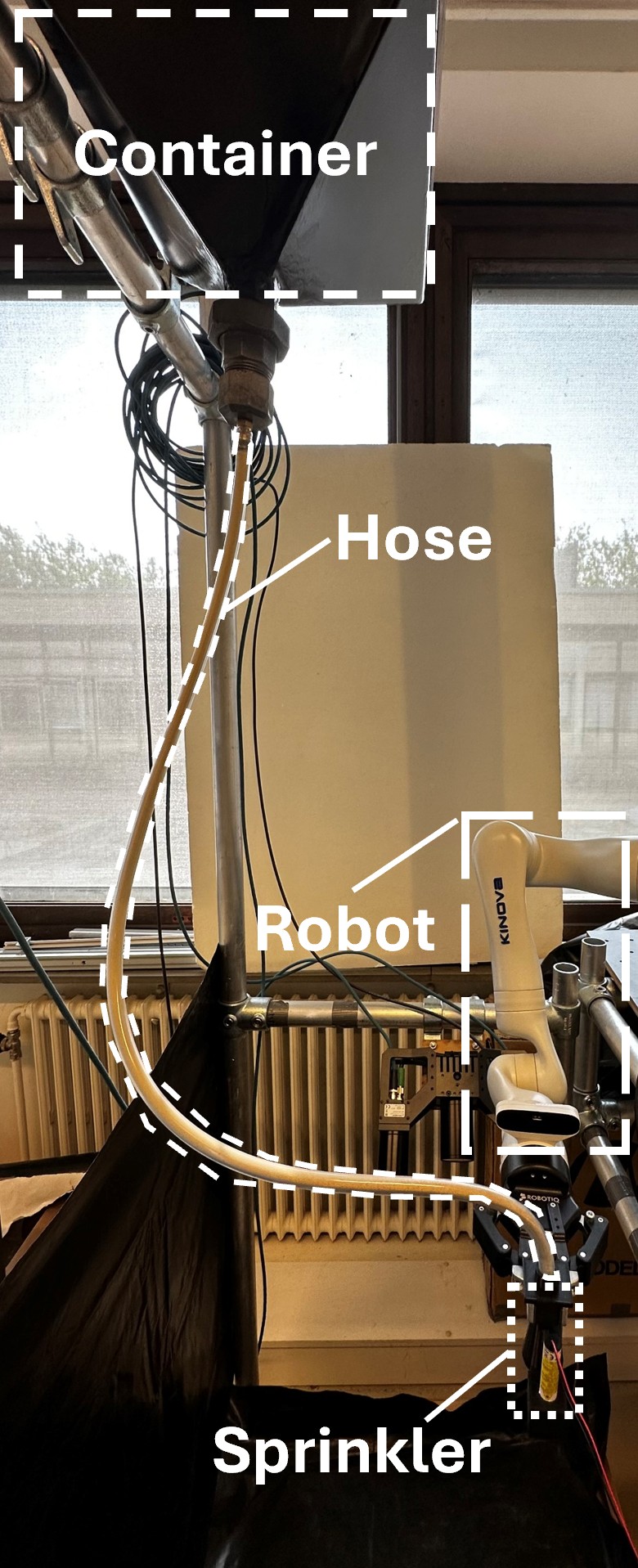}
            \caption{Side view}
            \label{fig:exp_setup_1}
        \end{subfigure}%
        \hfill
        \begin{subfigure}[t]{0.76\textwidth}
            \centering
            \includegraphics[width=\textwidth]{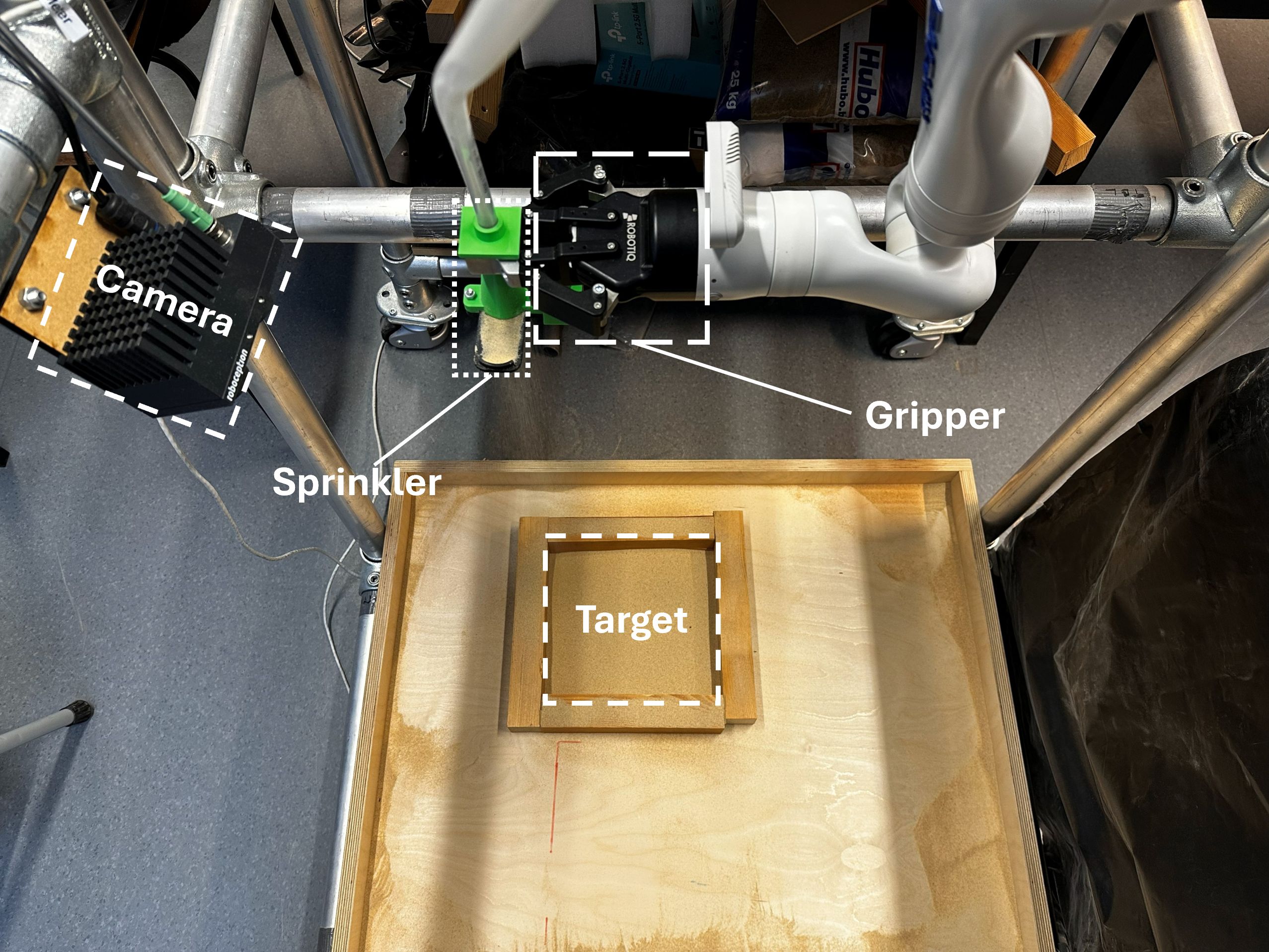}
            \caption{Top view}
            \label{fig:exp_setup_2}
        \end{subfigure}
        \caption{Sand-spraying testbed used for zero-shot sim-to-real transfer. (a) Side view: sand stored in an overhead container is transmitted through a hose to a sprinkler positioned over the target, generating plumes that induce visual occlusion. (b) Top view: the sprinkler is held by a robotic arm and guided using heightmaps of the target derived from an overhead stereo camera.}
        \label{fig:left-column}
    \end{minipage}%
\end{figure}

\begin{figure}[t]
  \centering
  \includegraphics[width=\columnwidth]{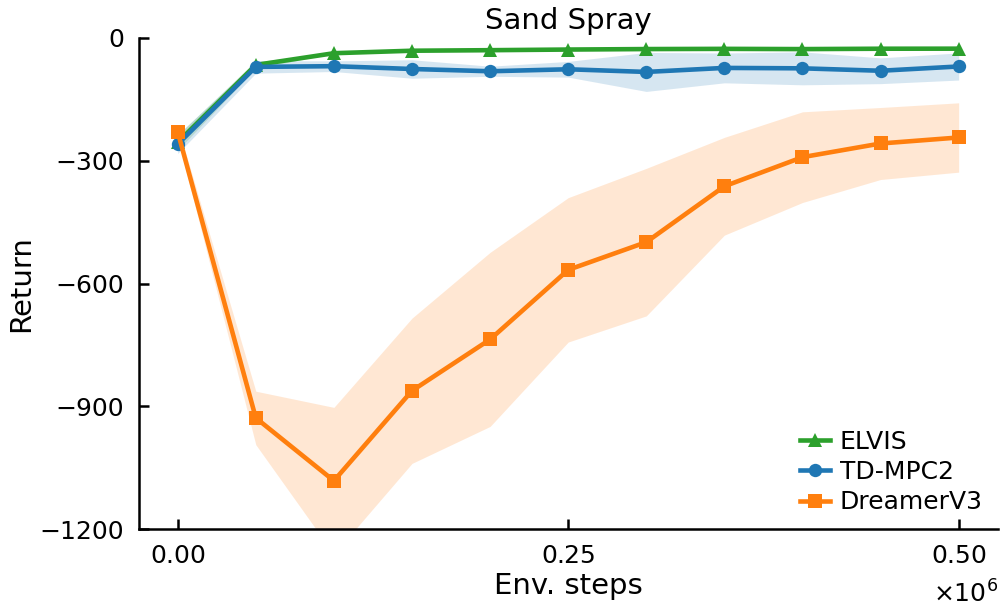}
  \caption{\textbf{Learning curves of simulated sand spray task.}
  Scores of ELVIS, TD-MPC2, and DreamerV3 on the simulated sand spray task, which shows the mean episodic return at each environment step. Shaded regions denote 95\% confidence intervals across 5 seeds. ELVIS significantly outperforms DreamerV3 while exhibiting more stable learning and reduced sensitivity to random seeds compared to TD-MPC2.}
  \label{fig:shotcrete-sim}
\end{figure}

\subsection{Sand Spray Sim-to-Real Zero Shot Transfer}
\label{sec:shotcrete}
We evaluate zero-shot sim-to-real transfer on a sand-spraying task that serves as a controlled proxy for real-world industrial operations in harsh, low-visibility settings, where airborne material and sensor noise induce severe occlusions. Following the experimental rationale of AREPO~\cite{du2025arepo}, a nozzle deposits granular material onto a target surface, and the control objective is to produce a uniform deposition profile despite partial observability and disturbances. Our simulator is a custom Gymnasium environment whose deposition dynamics are hand-designed from domain knowledge as detailed in AREPO, rather than built on top of a separate physics engine. In simulation, the agent observes simulated heightmaps and outputs planar nozzle-motion commands; after training, the learned policy is transferred without adaptation to the physical setup, where the robot is guided by real heightmaps reconstructed from an overhead stereo camera. 
Transfer does not rely on photorealistic rendering: the sim-to-real interface is primarily geometric through heightmaps, although a substantial gap remains due to dust, occlusion, sensor noise, and mismatch between simulated and real deposition dynamics. These factors reduce world-model prediction reliability, especially when observations are missing or corrupted over long-horizon rollouts. A detailed layout of the laboratory setup is shown in Fig.~\ref{fig:exp_setup_1} and Fig.~\ref{fig:exp_setup_2}.Performance is assessed using the surface-quality metrics defined in~\citep{du2025arepo}. Specifically, we report three surface-quality metrics: i) \textit{root-mean-square roughness} $\text{R}_{\text{rms}}$: the square root of the mean of the squares of the deviations of the surface height values from the mean surface height, ii) \textit{peak-to-valley roughness} $\text{R}_\text{t}$: the difference in height between the highest point and the lowest point on a surface, iii) \textit{waste volume ratio} $\text{r}_{\text{wv}}$: the ratio between the wasted volume and the desired volume to be fulfilled. The wasted volume is defined as the material volume that has been sprayed outside the target surface or that exceeds the target thickness.

\paragraph{Simulation learning performance}
Fig.~\ref{fig:shotcrete-sim} shows that model-based planning is a decisive advantage on this simulated sand spraying task: ELVIS and TD-MPC2 consistently outperform DreamerV3, which learns a latent world model but does not perform online MPC-style trajectory optimization at control time. The gap indicates that, in challenging visual control under extreme visual occlusions, explicitly optimizing action sequences through a predictive world model yields substantially stronger closed-loop behavior than purely policy-based action selection.


Beyond the benefits of longer-horizon planning, ELVIS offers a measurable advantage over TD-MPC2 in terms of learning robustness, as it exhibits more consistent and stable training with reduced sensitivity to random seeds. This improved stability is aligned with the DMC results and supports the claim that uncertainty awareness, together with an effective extension of the planning horizon, can improve data efficiency and make model-based control more reliable.

\begin{figure}[t]
  \centering
  \includegraphics[width=\columnwidth]{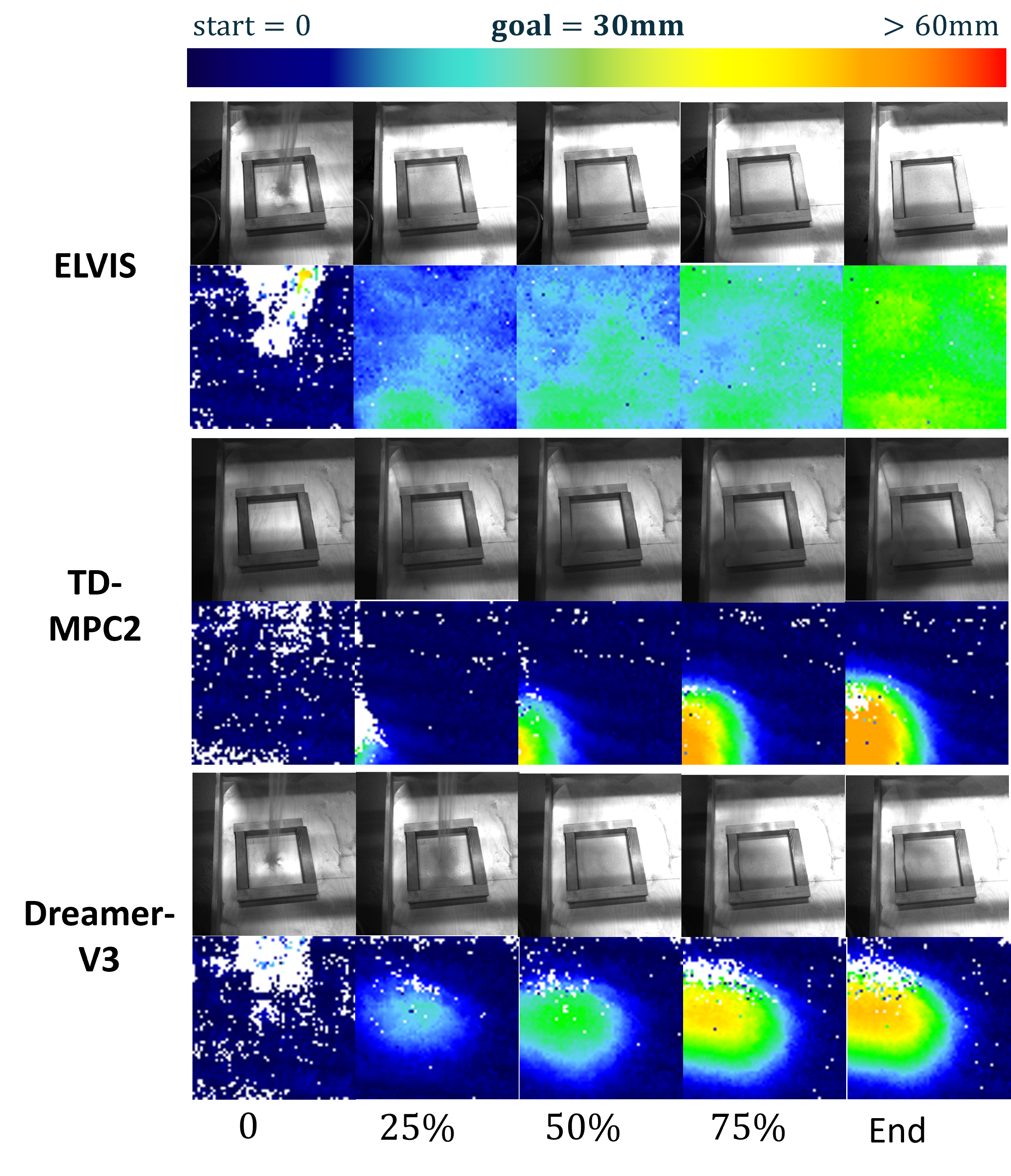}
  \caption{\textbf{Zero-shot evaluation with real-world sand spray task.}
For each method, the first row shows grayscale scene images for visualization only, while the policy itself acts on the corresponding heightmaps shown in the second row. In the sim-to-real experiment, ELVIS is most robust to partial observability caused by dust and sensory noise (unobservable white parts of the heightmaps) and achieves the best surface quality compared to other methods.}
  \label{fig:shotcrete-real}
\end{figure}
\paragraph{Real-world evaluation}
We transfer the trained policies to the laboratory sand-spraying testbed without adaptation and evaluate surface quality using the same metrics and protocol as~\citep{du2025arepo}. Each method is evaluated over 5 real-world hardware trials. Table~\ref{tab:sim2real} reveals a marked change in relative ranking from simulation to deployment: during simulation training we observe $\text{ELVIS}>\text{TD-MPC2}\gg\text{DreamerV3}$, whereas under zero-shot sim-to-real transfer the ordering becomes $\text{ELVIS}\gg\text{DreamerV3}>\text{TD-MPC2}$. This reversal highlights the importance of \emph{recurrent belief tracking} under severe real-world occlusion and sensor noise: methods that rely on naive frame stacking can struggle to maintain coherent control when observations become intermittently unreliable. Moreover, ELVIS’s advantage over DreamerV3 suggests that \emph{recurrent memory alone is not sufficient}: coupling belief-based control with \emph{uncertainty awareness} helps limit compounding model errors during long-horizon decision making and improves sim-to-real robustness.

These quantitative trends are reflected in qualitative behavior in Fig.~\ref{fig:shotcrete-real}. ELVIS produces the most uniform sand deposition throughout the entire spraying interval, yielding homogeneous coverage even under severe occlusions. In contrast, TD-MPC2 degrades sharply after transfer and collapses to repeatedly spraying a single corner, consistent with brittle planning under partial observability and unmodeled real-world perturbations. DreamerV3 transfers more reliably than TD-MPC2, supporting the benefit of recurrent memory, but its lack of online MPC refinement and explicit uncertainty handling limits its ability to recover from accumulated errors and to maintain consistent coverage. Together, these results demonstrate that ELVIS’s combination of long-horizon planning, recurrent belief, and uncertainty-aware truncation is essential for robust zero-shot real-world performance beyond simulation.

\begin{table}[t]
\caption{In zero-shot sim-to-real transfer to the laboratory testbed, ELVIS achieves the best surface quality compared to baselines without uncertainty awareness.}
\label{tab:sim2real}
\resizebox{\columnwidth}{!}{%
\fontsize{5}{6}\selectfont
\begin{tabular}{@{}llll@{}}
\hline
Metrics & \scalemath{0.9}{\text{R}_{\text{rms}}} (mm) & \scalemath{0.9}{\text{R}_{\text{t}}} (mm) & \scalemath{0.9}{\text{r}_{\text{wv}}} (\%) \\
\hline
ELVIS     & \textbf{2.2} $\pm$ \textbf{0.4} & \textbf{17.5} $\pm$ \textbf{0.5} & \textbf{6.3} $\pm$ \textbf{1.1} \\
TD-MPC2   & 16.3 $\pm$ 0.3                   & 69.6 $\pm$ 2.1                  & 47.2 $\pm$ 1.4 \\
DreamerV3 & 9.3 $\pm$ 0.4                   & 52.3 $\pm$ 1.1                  & 25.1 $\pm$ 0.7 \\
\hline
\end{tabular}%
}
\end{table}
\setlength{\textfloatsep}{4pt}

\section{Conclusion} 
\label{sec:conclusion}
We presented ELVIS, a memory-based, long-horizon, uncertainty-aware latent visual MPC framework for control under partial observability. Built on RSSM belief states, ELVIS combines (i) GMM-MPPI to preserve multi-modal long-horizon hypotheses and avoid mode collapse, and (ii) an ensemble-UCB-gated, time-varying $\lambda_t$ that softly truncates return propagation and links imagined actor-critic learning with online planning. Across standard DMControl vision benchmarks, ELVIS improves performance, and ablations support the roles of GMM-based long-horizon planning and uncertainty-aware truncation. ELVIS also achieves strong zero-shot sim-to-real transfer on a real-world sand-spraying task with severe visual occlusions, demonstrating robust deployment beyond simulation. The current study is limited to dense-reward continuous-control settings and incurs added compute cost from long-horizon latent MPC rollouts. Future work will study more efficient long-horizon world models and rollout-shortcut mechanisms, extend evaluation to sparse-reward manipulation, and develop tighter theory for mixture-based planning under model uncertainty.



\bibliographystyle{plainnat}
\bibliography{references}

\end{document}